# Zero-Shot Transfer of Force Map Estimation Across GelSight Mini Sensors

Julio Castaño-Amoros[1] and Pablo Gil[1,2]

[1]AUROVA Lab, Computer Science Research Institute, University of Alicante, San Vicente del Raspeig, 03690, Spain
[2]Department of Physics, Systems Engineering and Signal Theory, University of Alicante, San Vicente del Raspeig, 03690, Spain

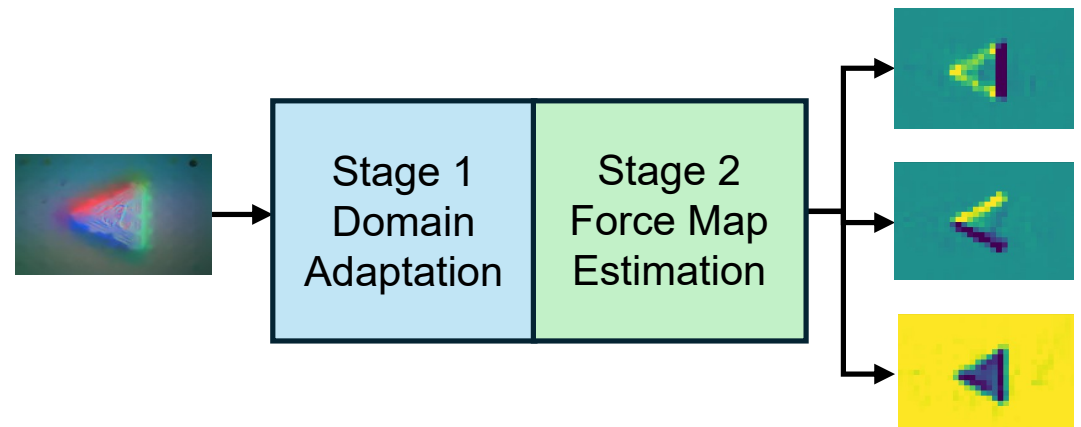

Abstract—Despite the rapid industrialization of the touch sensor manufacturing process, most of these sensors are still handmade in research laboratories. This complicates standardizing their performance, requiring the repetition of data collection and training models for each unit produced. To address this problem, this paper presents a method that can generalize the estimation of 3D force maps across different GelSight Mini sensor units, regardless of the sensor version. Specifically, the method consists of two stages: a domain adaptation stage, in which the input tactile image is reconstructed as a general tactile image using a UniT-based model; and a stage for estimating 3D force maps employing a U-Net network. Our proposal achieves promising results in both steps, such as an SSIM of $0.9338 \pm 0.0358$ in the image reconstruction phase and an $\text{MAE}_{\text{F}}$ of $1.1294 \pm 1.5934$ $(N)$ in the force estimation phase.

## I. INTRODUCTION

The manufacture of vision-based tactile sensors (VBTS) is not yet standardized, making the fabrication of identical sensor units a difficult process. Furthermore, this type of sensor is prone to breakage or malfunction due to its low-cost design. This fact forces repetition of the data collection and model training processes for each new sensor unit. The current approach to these challenges proposes a brute force strategy [1], [2], which involves the collection of large datasets (>500k samples) containing images of multiple VBTS. Therefore, it is possible to train large models based on transformers [3], [4], [5] to learn shared representations across different tactile images. However, this approach requires a huge amount of data and resources.

In contrast, this paper focuses on another approach based on learning intra-sensor tactile representations from smaller datasets via Self-Supervised Learning (SSL). Within this specific context, TacMAE [6] was proposed as a method to learn tactile representations from incomplete tactile data, employing a Masked Autoencoder (MAE) strategy and a dataset of 30k samples for tactile texture recognition. Another example is UniT [7], a model based on a VQGAN network [8] that was trained using 15k samples to reconstruct tactile images to learn tactile representations for various downstream tasks.

Although our work is based on UniT, it has key differences that enhance performance. Our method can reconstruct tactile images without markers, unlike [7], and also generates reconstructed images with a higher level of detail. Our proposal was trained with a larger dataset of approximately 115k markerless images obtained from a single GelSight Mini sensor and a larger number of objects (55) while UniT used a small dataset of tactile images with markers from a single object.

Furthermore, while UniT was evaluated with two unseen Gelsight Mini sensors that generate tactile images very similar in appearance to those used in the training dataset, we have used several Gelsight Mini sensors presenting different color backgrounds as examples of intra-sensory variation. Note that the original UniT model was tested on various downstream tasks by freezing the encoder and training a different decoder for each task. However, the latent space learned by UniT may be insufficient for certain tasks. In this regard, our work also evaluates the potential of UniT as a domain adaptation model to transform the input tactile image from different GelSight Mini units into a generalist tactile reconstructed image, which can then be used as input to a task-specific model. Specifically, we have evaluated the zero-shot generalization capabilities of the proposed method, which comprises a UniT-based model and a U-Net network [9] for the task of estimating the force map.

Corresponding author: J. Castaño-Amorós (e-mail: julio.ca@ua.es).

## II. FORCE MAP ESTIMATION ON GELSIGHT VBTS

### A. The proposed method

In this paper, we propose a method that generalizes the estimation of distributed forces across tactile images from different GelSight Mini sensor units even if it has only been trained with images from one sensor unit (Fig. 1). The proposed method consists of two stages: a first VQGAN-type neural network that acts as a domain adaptation method between the different GelSight sensors generating a reconstructed intermediate image ($I_r$, $I_d$), and a second U-Net neural network to estimate from $I_d$ the force maps ($I_{fx}$, $I_{fy}$, $I_{fz}$) distributed on the sensor's surface as well as the magnitude of the total force.

**Intra-sensor domain adaptation.** This stage is based on a VQGAN architecture, which consists of a VQVAE model [10] and a discriminator. The VQVAE comprises a Convolutional Neural Network (CNN) encoder and decoder, with a quantization layer between them. This layer acts as a codebook, transforming the latent space from a continuous to a discrete distribution by mapping each latent value to its closest codebook value. Consequently, the model only learns a fixed library of tactile features that it can then generalize to other cases. Besides, the discriminator is a patch-based CNN that is only used during training to learn high-quality reconstructions. In summary, a tactile image $I_i$ obtained from any sensor unit can be transformed into a general representation image $I_r$.

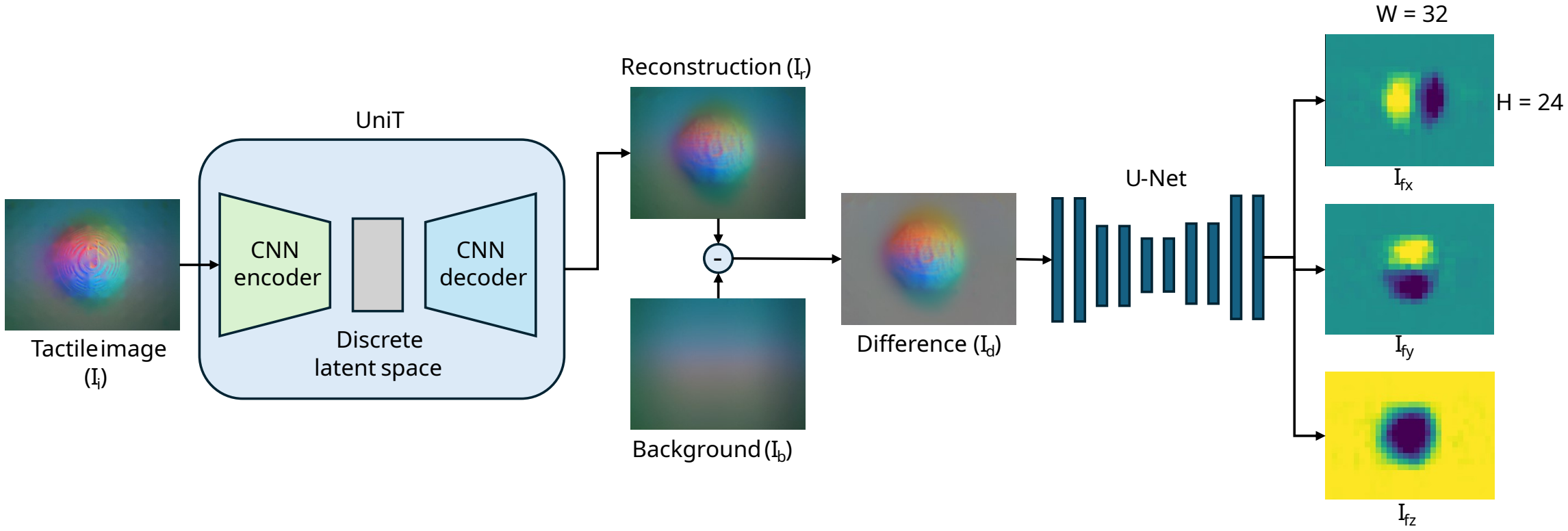

Fig. 1. Overview of the proposed method to generalize the estimation of force distributions across different units of GelSight Mini sensors.

**Force map estimation.** In the second stage, U-Net receives as input a difference tactile image $I_d$, obtained by subtracting the reconstruction of the input image $I_r$ from the reconstruction of the background image $I_b$. This architecture consists of an encoder-decoder structure. The encoder contracts the feature map, and the decoder expands it. In the encoder, convolutional layers, ReLU functions, and max-pooling layers are applied iteratively to extract features at different levels of detail. The decoder combines subsampling operations with the feature concatenation from the encoder, and finally applies two convolutional layers to obtain three images representing the force maps ($I_{fx}$, $I_{fy}$, $I_{fz}$). The forces ($F_x$, $F_y$, $F_z$) are then calculated from these force maps as $\sum_{k=1}^{W \cdot H} I_{fj}(k)$, where j={x,y,z} and $W \cdot H$ is the size of the force map. Finally, the magnitude of the force is obtained as $F = \sqrt{{F_x}^2 + {F_y}^2 + {F_z}^2}$.

## B. Data collection

Two distinct datasets are created. One is used for force reconstruction and the other for force map estimation.

**Reconstruction dataset.** The data collection process for tactile image reconstruction involves touching 55 different objects multiple times with a single GelSight Mini sensor. A human operator generates touches in different poses for two minutes, collecting approximately 2,300 tactile images per object (115,042 in total). Five objects were selected from each of the following categories: common (e.g., phone case), garments (e.g., jeans), deformable/semi-deformable (e.g., bottle), multiple contacts (e.g., LEGO plane), stationery (e.g., pen), toys (e.g., reptile), sports (e.g., tennis ball), relief (e.g., magnet), paper (e.g., kitchen paper), and 3D-printed (e.g., square).

**Force dataset.** The dataset used for the force estimation task is based on [11], in which the authors collected a force distribution dataset by generating random contact configurations between different GelSight Mini sensors with markers and 24 3D-printed indenters. This dataset comprises a total of 18,679 samples, including the respective GelSight Mini images with markers and force map labels obtained using a Finite Element Analysis (FEA) from [11]. The authors of [11] created 25 different sensor configurations by swapping the gels in five GelSight Mini units, generating variations in sensor characteristics such as, for example, the image background, the gel thickness or stiffness. Consequently, we have not been able to measure these changes in the sensor properties. The ground truth forces range from 0 to 40 ($N$) in the z-axis and from 0 to 5 ($N$) in the x-y axes.

For our work, we needed to remove the markers in the tactile images within that dataset. Therefore, we developed an algorithm to remove the markers from the tactile images to generate markerless images $I_i$ (Fig. 2). Although the use of markers minimizes force errors, it can result in a poorer perception of other object characteristics due to the occlusions that markers cause. Specifically, the algorithm works as follows. First, the $RGB_m$ images with markers (Fig. 2a and 2e) are reconstructed to obtain a shared background color (Fig. 2b and 2f). Second, a threshold operation is applied to get the mask of the markers of the reconstructed images. Next, the TELEA [14] inpainting algorithm is employed to remove the markers using the mask and fill in the surrounding area in the original images (Fig. 2c and 2g). Finally, the images are reconstructed using our UniT-based model to remove or smooth artifacts, resulting in RGB tactile images denoted as $I_i$ that share the background colors across different sensors (Fig. 2d and 2h). After applying the dot removal algorithm, 504 images were discarded because it failed to remove most of the black dots. This left a total of 18,175 markerless images from the original dataset with markers, as described in [11].

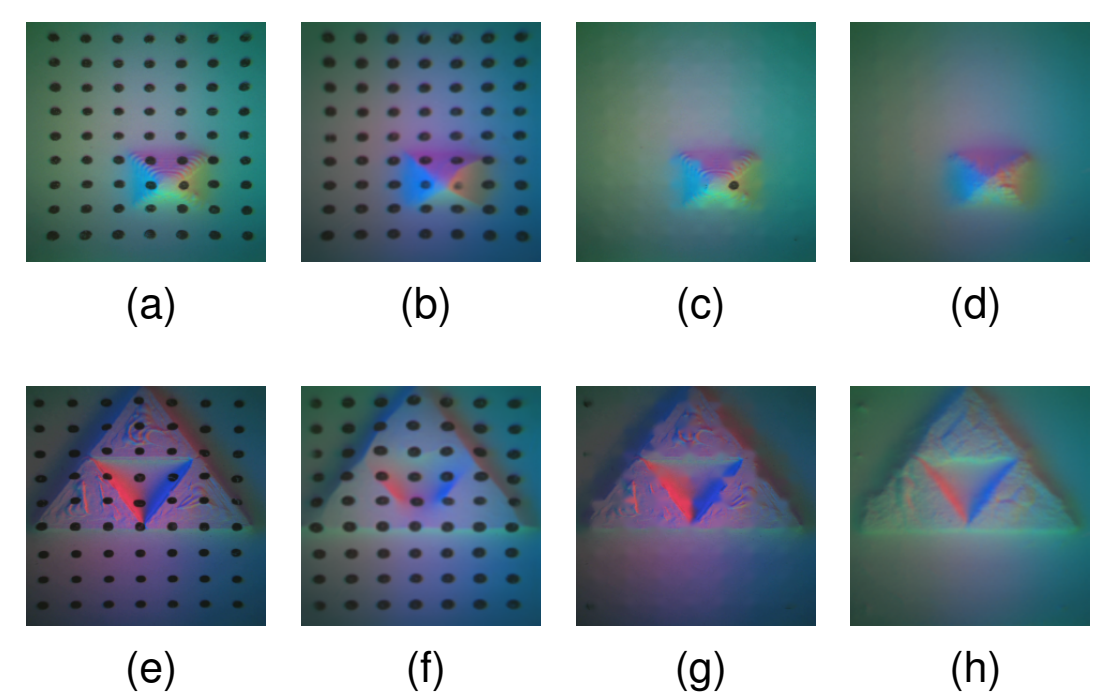

Fig. 2. Examples of transformation of tactile images with markers, $RGB_m$, to tactile images without markers, $I_i$.

## C. Training

The proposed method required two training phases: a reconstruction phase and a force estimation phase.

In the reconstruction phase, the UniT architecture was adapted to create three models by modifying the dimensions of the feature maps used in the convolutional layers of the encoder and the decoder. These models were named *small*, *medium* and *large* and comprised 25.95M, 35.66M and 79.81M parameters, respectively. These models were trained using the reconstruction dataset described in the previous section, organized into training and testing sets at a ratio of 90 : 10. All models were trained using the following hyperparameters: a batch size of 32, a learning rate of $4.5 \times 10^{-6}$, VQLPIPSWithDiscriminator[1] as the loss function, and 35 epochs.

In the force estimation phase, 19 indenters were used for the training and validation sets, and 5 new indenters, not seen before, were used for the test set. The data were divided into three sets of images: 15,447 for training, 908 for validation and 1,820 for testing. During training, the U-Net model learned to estimate force distributions from reconstructed tactile images with the same background (Fig. 2d and 2h). All trainings were carried out using the following hyperparameters: a batch size of 64, a learning rate of $6 \times 10^{-4}$, Mean Squared Error (MSE) as the loss function, and 100 epochs. Data augmentation techniques, such as adding Gaussian noise and adjusting image brightness, contrast, saturation, and hue, were also applied.

## III. RESULTS

### A. Tactile image reconstruction

The performance of each implemented model (*small*, *medium*, *large*) was compared with the state-of-the-art methods, MAE [12] and pix2pix [13]. All methods were trained using the same dataset described in Section II-B. To evaluate, we used the test set containing 11,503 images, with the Structural Similarity Index Metric (SSIM) and the Fréchet Inception Distance (FID) as the performance metrics.

On the one hand, the first two columns of Table 1 show that the *small*, *medium*, *large* and pix2pix methods achieved higher SSIM and lower FID values than those obtained by MAE, being *medium* the best model and *small* the second best. This is because the MAE model is designed for global context learning rather than for high-quality image reconstruction. Consequently, unlike pix2pix and VQGAN models, the MAE architecture does not include a discriminator network, resulting in blurrier reconstructions and higher FID values.

TABLE 1. Comparison of reconstruction results for intra-sensor domain adaptation in terms of the mean and standard deviation of SSIM and FID, without using and using the augmented test set.

| Method | SSIM ↑ | FID ↓ | SSIM ↑ | FID ↓ |
|---|---|---|---|---|
| *small* | 0.9289 ± 0.0391 | **17.8018** | 0.9043 ± 0.0468 | 28.0208 |
| ***medium*** | **0.9338 ± 0.0358** | 18.2523 | **0.9168 ± 0.0404** | **27.5062** |
| *large* | 0.9295 ± 0.0372 | 20.8332 | 0.9071 ± 0.0463 | 29.5612 |
| MAE [12] | 0.8313 ± 0.0462 | 60.6758 | 0.4754 ± 0.0987 | 170.5869 |
| pix2pix [13] | 0.9075 ± 0.0033 | 26.9036 | 0.7610 ± 0.0034 | 188.4666 |

On the other hand, in the last two columns of Table 1, we also show the generalization capability of each model for the reconstruction task, using the augmented tactile image dataset. Specifically, the brightness, contrast, saturation and hue of the tactile images were modified by sampling four values from a uniform distribution within the range $[0.1, 0.3]$. This allows us to determine if the trained models can apply the learned features to different lighting conditions.

As can be seen, the three models adapted from UniT once again outperform other state-of-the-art methods, being *medium* the best model. Their SSIM and FID values worsened by no more than 3 and 11 units, respectively. In contrast, the performance of the MAE and pix2pix methods deteriorated significantly in both metrics. The superior generalization of the proposed methods is thanks to the use of a codebook and a discrete latent space inherited from UniT. Rather than memorizing tactile features, these models learn to map them into a fixed database of tactile features. Consequently, they do not hallucinate when processing tactile images that differ from those in the training set. Instead, these models identify the features in the codebook that are the most similar. Figure 3 illustrates this fact with two examples of tactile image reconstruction. As can be seen, the reconstructions from the MAE model reflect the poor results obtained in the second-to-last row of Table 1. Unlike MAE, pix2pix model generated higher-quality reconstructions but these remained within the same domain as the augmented input images, since each $\mathrm{I_r}$ is more similar to the augmented image than to the original $\mathrm{I_i}$. In contrast, our *medium* model achieved a better balance between quality reconstruction and generalization to the original tactile image, as proves the similarity of $\mathrm{I_r}$ shown in Fig. 3c and 3h with $\mathrm{I_i}$ shown in Fig. 3a and 3f respectively.

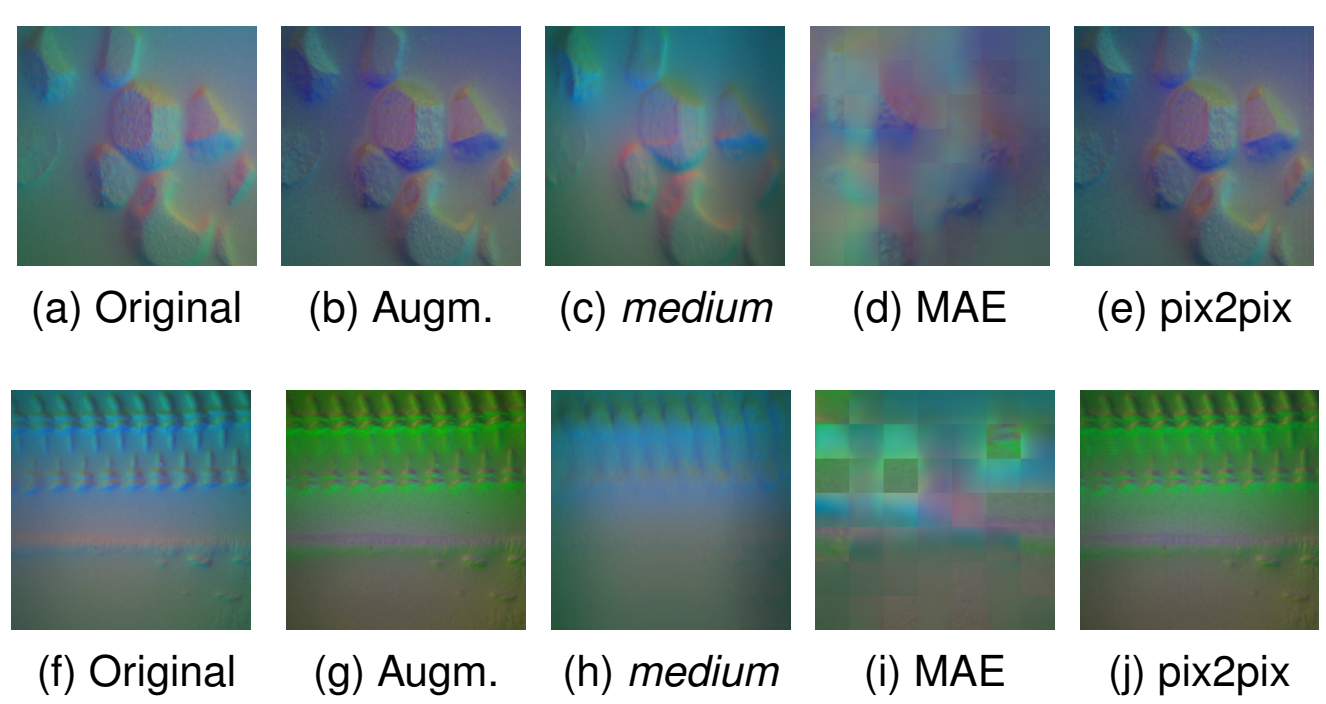

Fig. 3. Examples of the results of tactile image reconstruction using the augmented test set for each of the models.

### B. Tactile force map estimation

In this experiment, we evaluate the force map estimation of our method using three tactile representations: ($\mathrm{I_r}$) is the tactile image obtained after reconstructing the input tactile image without markers denoted as $\mathrm{I_i}$. ($\mathrm{I_d}$) is the tactile image obtained by subtracting the reconstructed image $\mathrm{I_r}$ and a background image denoted as $\mathrm{I_b}$. Finally, ($\mathrm{RGB_m}$) is an image with markers from the dataset presented in [11], which is used as the performance reference for this task.

To evaluate the performance, we calculate the Mean Absolute Error of the predicted force (F) and its ground truth for each axis, as well as for the force magnitude ($\mathrm{MAE_F}$) as shown in Table 2.

TABLE 2. Force errors obtained in terms of the $\mathrm{MAE_F}$ and expressed in Newtons (N). Note that the $\mathrm{RGB_m}$ is the reference for the task.

| Input | $\mathrm{MAE_{Fx}}$ | $\mathrm{MAE_{Fy}}$ | $\mathrm{MAE_{Fz}}$ | $\mathrm{MAE_F}$ |
|---|---|---|---|---|
| $\mathrm{RGB_m}$ | 0.2460 ± 0.3944 | 0.1653 ± 0.2545 | 0.7234 ± 0.9666 | 0.7250 ± 0.9668 |
| $\mathrm{I_r}$ | 0.2888 ± 0.4729 | 0.2539 ± 0.4097 | 1.1926 ± 1.6470 | 1.1949 ± 1.6476 |
| $\mathbf{I_d}$ | **0.2978 ± 0.4922** | **0.2534 ± 0.3966** | **1.1256 ± 1.5937** | **1.1294 ± 1.5934** |

As is widely known, images with markers offer better performance than those without markers for the force estimation task. The aim is to prove whether the markerless reconstructed images from our intra-sensor domain adaption can achieve similar force estimation performance. The results prove that $I_r$ and $I_d$ achieved average $MAE_F$ values that were $0.4699 \pm 0.6808$ $(N)$ and $0.4044 \pm 0.6266$ $(N)$ higher on average, which are not significant differences. $Id_d$ performs better than $Ir_r$ because the subtraction removes the background and highlights the tactile contact features, which are more related to the applied force.

[1] VQLPIPSWithDiscriminator stands for Vector Quantized Learned Perceptual Image Patch Similarity with Discriminator.

In addition, the zero-shot generalization capabilities of the complete method were tested using images from the test set of the force map dataset, which were obtained before a shared background was generated (Fig. 2c and 2g). To do this, our method was accelerated using CUDA Graph, achieving an inference time of $19.7809 \pm 0.9886$ $(ms)$ over 500 runs on a 3060 RTX. This is sufficient for real-time force map estimation, as the GelSight Mini sensor runs at 25 $(Hz)$. Fig. 4 shows an example of force map prediction obtained by our method. The first stage reconstructs the original image $I_i$ into a shared-background image $I_r$ and the second stage estimates the force maps from the difference image $I_d$. It is important to note that the proposed model can estimate force maps from tactile images despite GelSight Mini sensors exhibit intra-sensor variance.

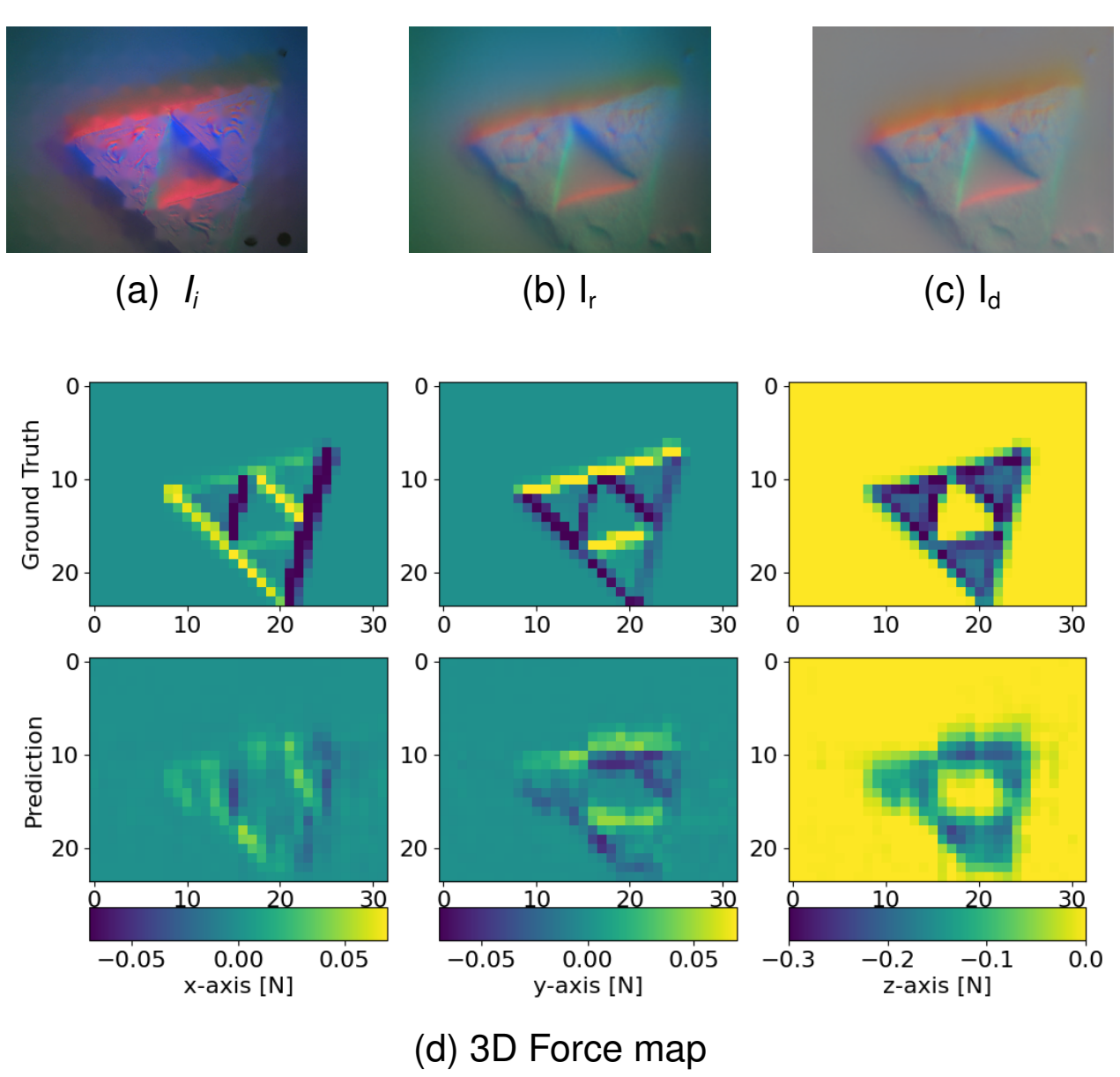

Fig. 4. Example of force map estimation carried out by the proposed method using a triangular indenter.

## IV. CONCLUSION

This paper introduces a method for estimating 3D force distributions from GelSight Mini sensors which exhibit variances that cause different colored backgrounds. The strength of our approach is that the proposed model can be generalized to other sensors, despite having only been trained with images from a single sensor. This demonstrates our model's high generalization capabilities and improves upon the original UniT model by achieving high-quality, generalizable reconstructions. Furthermore, our markerless approach yielded $MAE_F$ results comparable to those obtained using tactile images with markers. However, our method had difficulty predicting very low or very high normal forces (near 0 $(N)$ or 40 $(N)$ in the z-axis) as well as shear forces greater than 5 N in the x-y plane. Besides, our model struggles to predict the contact geometries correctly. Future work will focus on adding a loss function to help the U-Net model preserve contact geometry during training.

## ACKNOWLEDGMENT

This work was supported by the Interreg-VI Sudoe and European Regional Development Funds through the REMAIN Project under Grant S1/1.1/E0111 and by the University of Alicante under Grant UAFPU21-26.